\documentclass{tlp}

\usepackage{multirow}
\usepackage{natbib}
\usepackage{colonequals}
\usepackage{bm}
\usepackage{booktabs}
\usepackage{listings}
\usepackage{subcaption}
\usepackage{xcolor}
\usepackage{hyperref}

\begin{document}

\lefttitle{Rader and Russo}

\jnlPage{1}{16}
\jnlDoiYr{2026}
\doival{}

\title[Accelerating NeurASP with vectorization and caching]{Accelerating NeurASP with vectorization and caching}

\begin{authgrp}
\author{\gn{Alexander Philipp} \sn{Rader} and \gn{Alessandra} \sn{Russo}}
\affiliation{Imperial College London \\ \emails{apr20@ic.ac.uk}{a.russo@ic.ac.uk}}
\end{authgrp}


\maketitle

\begin{abstract}
Neurosymbolic AI combines neural networks with symbolic programs to create robust and explainable predictions.
One such framework is NeurASP, which trains a neural network to predict concepts and reasons over them using rules written in answer set programming (ASP) to solve downstream tasks.
Crucially, labels are only provided for the downstream prediction produced by the symbolic rules, not for the latent concepts themselves.
Backpropagation through the non-differentiable ASP component requires expensive probability and gradient calculations, which has hindered scalability to more sophisticated tasks.
In this paper, we address the current limitations of NeurASP by improving its computational performance through vectorization, batch processing and caching of intermediate computations during training.
We compare computation speeds between the original and our new implementation of NeurASP and report speedups of multiple orders of magnitude for larger tasks.
To this end, we propose a new dataset of difficult tasks involving playing cards, which we use to test the capabilities of NeurASP's enhanced learning function.
\end{abstract}


\section{Introduction}

In neurosymbolic artificial intelligence (AI), neural networks are combined with symbolic languages to harness the strengths of both.
Neural networks excel at handling real-world data and learning autonomously using backpropagation, while symbolic specifications are explainable and capable of robust reasoning with safety guarantees.
Their combination has the potential to create frameworks that are transparent, trustworthy and capable of solving real-world problems~\citep{garcez2023third}.

One symbolic language is answer set programming (ASP), which belongs to the family of logic programming.
ASP consists of rules that model complex problems and uses automated reasoning to find solutions~\citep{lifschitz2019asp}.
Thanks to its expressivity, it is a popular choice for neurosymbolic systems.
There are a wide variety of frameworks that combine neural networks and ASP in different ways:
Some hard-code the ASP component~\citep{aspis2022embed2sym,skryagin2024scalable_slash,geh2024dpasp}, others use pre-trained networks, such as large language models, to generate ASP~\citep{cunnington2024foundation_reasoning,wang2024dspy,kalyanpur2024llm-arc}.
The framework NeurASP~\citep{yang2020neurasp} belongs to the former category.
It uses a neural network to process input data and predict latent concepts.
A hard-coded ASP specification takes in these concepts and arrives at a downstream prediction.

Typically in neurosymbolic reasoning, no labels for the latent concepts are available, only for the downstream labels.
In order to backpropagate through the non-differentiable ASP component, NeurASP makes use of semantic loss~\citep{xu2017semantic_loss}.
It first generates all possible latent combinations, called answer sets, that lead to a correct downstream label.
Next, it assigns a probability to each answer set and calculates gradients based on them.
All three of these computations are expensive and therefore present a bottleneck for NeurASP\@.
In a number of existing works, NeurASP has been shown to time out when applied to more challenging datasets~\citep{aspis2022embed2sym,aspis2024embed2rule,cunnington2024foundation_reasoning}.

In this paper, we address these limitations and present improvements to every step of the NeurASP algorithm.
Our enhanced framework substantially accelerates learning performance, improving scalability to more complex problems.
It employs caching of answer sets, ensuring that they are calculated only once per downstream label, and uses vectorized operations in place of naive loops for probability and gradient calculations.
To improve usability, we have included support for data loaders and batch processing, comprehensive logging, seeding for reproducible results and validation sets for hyperparameter tuning.

We run performance tests on synthetic data and a variety of benchmark tasks.
Since existing neurosymbolic datasets contain simple perception inputs with few latent concepts, we introduce Card arithmetic, a new set of tasks with a higher complexity using playing card images.
The results show that our improved version achieves speed gains of multiple orders of magnitude compared to NeurASP and SLASH~\citep{skryagin2024scalable_slash}, a recent extension of NeurASP\@.
Moreover, we demonstrate that NeurASP can successfully solve Card arithmetic tasks, achieving high accuracies even with dozens of concepts and tens of thousands of answer sets.
It outperforms Embed2Sym~\citep{aspis2022embed2sym}, another ASP-based learning framework, which times out.

Our paper is structured as follows: Section~\ref{sec:related} provides an overview of neurosymbolic frameworks similar to NeurASP\@.
Section~\ref{sec:prelim} introduces the basics of ASP and NeurASP\@.
Section~\ref{sec:improvements} outlines the technical enhancements we have made for our improved framework.
Section~\ref{sec:performance} showcases computation time speedups on synthetic data.
Section~\ref{sec:dataset} describes benchmark tasks and introduces the Card arithmetic dataset.
Section~\ref{sec:experiments} presents the accuracy results of NeurASP and Embed2Sym.
Section~\ref{sec:conclusion} provides concluding remarks and open research directions.

The improved implementation, along with the new Card arithmetic dataset, can be found at \url{https://github.com/azreasoners/NeurASP}.

\section{Related work}
\label{sec:related}

In this section, we describe key neurosymbolic reasoning frameworks in the field.
We start with NeurASP~\citep{yang2020neurasp}, as it is the framework we improve in this paper.
NeurASP integrates neural networks with answer set programs such that the network predicts latent concepts and the program calculates a downstream prediction using these concepts.
It employs semantic loss~\citep{xu2017semantic_loss} to propagate the learning signal from the downstream labels to the network, since latent labels are not available.
The framework SLASH extends NeurASP by replacing the neural component with probabilistic circuits, which can estimate more complex probability distributions for latent concepts~\citep{skryagin2022slash}.
A further extension improves the scalability of SLASH by pruning insignificant latent concepts during learning~\citep{skryagin2024scalable_slash}.
We will compare our improvements with both of these frameworks, as they have the core architecture in common.

NeurASP and SLASH belong to the category of neurosymbolic systems that train a neural network with a hard-coded answer set program.
Another framework in this category is dPASP, which enables neural predicates to be interval-valued and facts to be annotated with probabilities~\citep{geh2024dpasp}.
A different approach is used by Embed2Sym, which first trains a neural network end-to-end on downstream labels and then utilizes clustering of the embedding space to extract latent concepts.
The ASP program is used to match clusters with concepts in a way that maximizes downstream performance~\citep{aspis2022embed2sym}.

Stepping beyond ASP, there are other frameworks that integrate neural and symbolic knowledge in a similar manner.
DeepProbLog~\citep{manhaeve2021deepproblog} extends the ProbLog language with support for neural predictions, which act as probabilistic facts.
MetaABD~\citep{wang-zhou2021abductive_knowledge} goes a step further and trains a neural network while learning a definite logic program at the same time.

\section{Preliminaries}
\label{sec:prelim}

Here we introduce the basic concepts used throughout the paper.
We outline the basics of the logical language, ASP, and detail how NeurASP works.

\subsection{Answer set programming}

We give a succinct account of all the concepts in ASP that are relevant to this paper, adapted from \citet{law2019las}.
For a thorough description, please refer to \citet{lifschitz2019asp}.

An answer set program consists of a set of rules.
We typically deal with normal rules, which are made up of atoms in the following manner:
\[\mathtt{h\colonminus b_1,\ldots,b_m,not\ b_{m+1},\ldots not\ b_n}.\]
where the atom $\mathtt{h}$ represents the head and $\mathtt{b_1,\ldots,b_m,not\ b_{m+1},not\ b_n}$ make up the body of the rule.
The symbol $\mathtt{not}$ is called negation as failure and is true if its corresponding atom cannot be derived from the rules in the program.

A solution of an answer set program $P$ is a set of ground atoms $I$, i.e.\ atoms that do not contain variables, and is defined in terms of the reduct of $P$.
$I$ is a subset of the Herbrand base $HB_P$, which is the set of ground atoms that can be formed from the predicates and constants in $P$.
Given such an interpretation $I\subseteq HB_P$, the reduct $P^I$ is constructed as follows:
First, the grounding of $P$ is constructed.
Second, all rules are removed from the grounding that contain an atom of the form $\mathtt{not\ b}$, where $b\in I$.
Third, all remaining negation as failure atoms are removed.
This results in a definite program, and $I$ is a model if it makes every rule in it true.
If $I$ is the smallest possible model, it is minimal.
$I$ is a solution of $P$ if it is a minimal model of the reduct $P^I$ and is called an answer set or stable model of $P$.
A program might have multiple answer sets.

The language provides many constructs to facilitate modelling, two of which are relevant for NeurASP\@.
A choice rule has the form
\[\mathtt{1\{h_1,\ldots,h_k\}1\colonminus b_1,\ldots,b_m,not\ b_{m+1},\ldots,not\ b_n}.\]
Whenever the body holds, exactly one of the head atoms has to hold.
Since there is a choice between head atoms, each decision can create an additional answer set.
A weak constraint provides an ordering of answer sets and has the form
\[\mathtt{\colonsim b_1,\ldots,b_m,not\ b_{m+1},\ldots,not\ b_n}.[w@l,t_1,\ldots,t_o]\]
$\mathtt{w}$ and $\mathtt{l}$ specify the weight and optional priority level of each constraint, while $t_1,\ldots,t_o$ are identifiers.
Optimal answer sets minimize the sum of the weights of all weak constraints whose bodies are true, starting at the highest priority level.

\subsection{NeurASP}
We provide an overview of the structure of NeurASP and its learning mechanism.
For a detailed explanation, please refer to the original paper from \citet{yang2020neurasp}.

NeurASP combines neural networks with ASP in a cascading fashion:
A neural network takes in data and predicts latent concepts, which in turn are integrated into an answer set program and processed by a solver to create a downstream label.

\begin{figure}
\includegraphics[width=\textwidth]{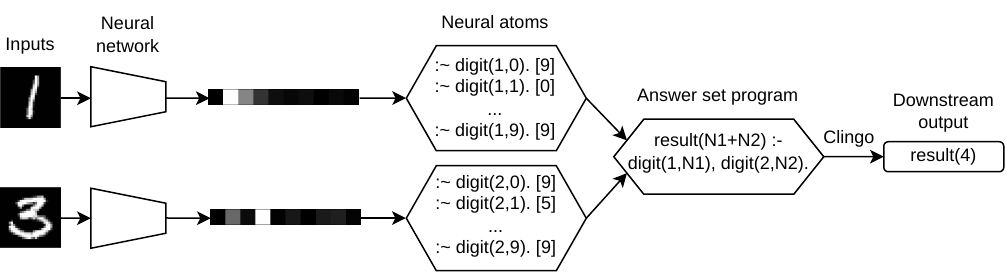}
\caption{The inference procedure of NeurASP exemplified with an MNIST addition example.}
\label{fig:neurasp_inference}
\end{figure}

We illustrate the workings of NeurASP with an example task:
In MNIST addition, each datapoint consists of two MNIST~\citep{deng2012mnist} images depicting handwritten digits and a downstream label representing their sum.
Crucially, no latent labels of the digits themselves are available, only of their sum.

The inference procedure of NeurASP is shown in Figure~\ref{fig:neurasp_inference}.
First, the framework runs each image through a neural network, yielding prediction vectors.
Each entry in the vector represents a symbolic concept, known as a neural atom, and is encoded as a weak constraint.
Entries with higher probabilities are given lower weights and are therefore preferred in the minimization task.
We omit the identifiers for the sake of conciseness.
The weak constraints are added to the hard-coded answer set program, which in this case calculates the sum of the two neural atoms.
The answer set solver Clingo~\citep{gebser2017clingo} then computes the optimal answer set of the program and the result is extracted from it.

While inference is straightforward, training the neural network to predict the correct neural atoms is challenging, as no latent labels are provided.
Instead, the learning signal has to be propagated from the downstream label through the non-differentiable answer set program to the neural network.
NeurASP solves this challenge by utilizing semantic loss, which rewards the neural network for all latent predictions that lead to a correct downstream prediction~\citep{xu2017semantic_loss}.
The authors adapt semantic loss to ASP by computing all possible answer sets that yield a correct downstream prediction and training the neural network to predict concepts that are present in these answer sets.
This process involves three major steps:

\begin{enumerate}
    \item \textbf{Answer set computation:} Given a downstream label, all possible answer sets are computed that predict that label.
      Each answer set contains a unique combination of neural atoms.
    \item \textbf{Probability calculation:} Each answer set is given a probability based on the neural network confidences of neural atoms within it.
    \item \textbf{Gradient calculation:} The gradient of the semantic loss with regard to the neural network outputs is computed and propagated through all parameters.
\end{enumerate}

Before we explore each step in more detail, we introduce the necessary notation.
Let $\Pi^{asp}$ denote the hard-coded ASP program and $\Pi^{nn}$ be the set of neural atoms.
A neural atom $c$ can take on a value $v\in\{v_1,\ldots,v_n\}$.
The probability $P(c=v)$ is given by the corresponding entry of the output vector of a neural network $N$ that has processed input $x$.
$N$ has parameters $\bm{\theta}$ that are learnable.
The downstream label is an observation $O$ and a set of neural atom values form an interpretation $I$.

\subsubsection{Answer set computation}

\begin{figure}
\includegraphics[width=\textwidth]{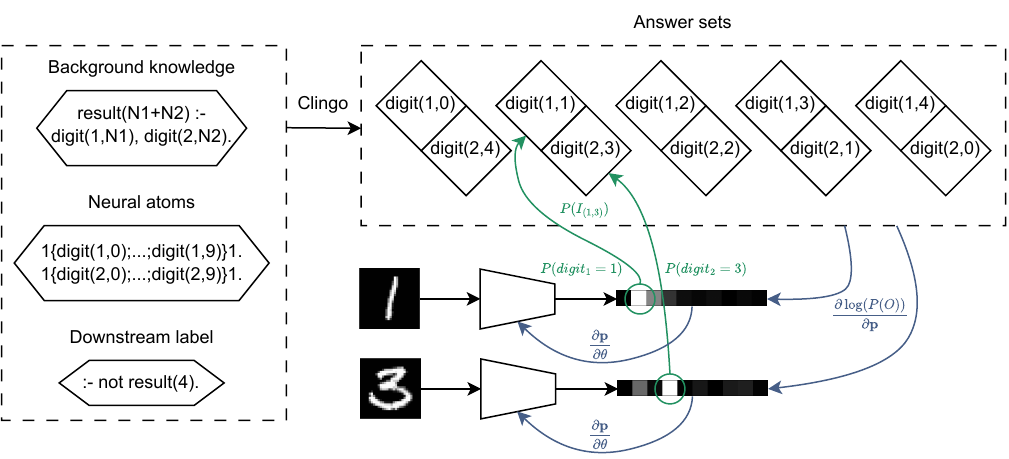}
\caption{The learning procedure of NeurASP exemplified with an MNIST addition example.}
\label{fig:neurasp_learning}
\end{figure}

The program $\Pi$ is composed of $\Pi^{asp}$, the downstream label $O$ and the neural atoms $\Pi^{nn}$.
$O$ is encoded as a constraint, ensuring that each answer set must prove $O$.
Each neural atom $c_i$ is encoded as a choice rule $\mathtt{1\{c(i,v_1);\ldots;c(i,v_n)\}1}$, which instructs Clingo to choose exactly one value for the atom per answer set.

Figure~\ref{fig:neurasp_learning} illustrates the answer set computation for an example in MNIST addition.
In this case, Clingo computes five valid answer sets with neural atoms that sum to 4: $\{(0,4),(1,3),(2,2),(3,1),(4,0)\}$.

\subsubsection{Probability calculation}

The probability of a set of neural atom values $I$ is determined by its neural network confidences:

\begin{equation}
P(I)=\prod_{c=v\in I}P(c=v)
\label{eq:prob}
\end{equation}

Answer sets that contain neural atom values with higher confidences thus attain higher probabilities.
Figure~\ref{fig:neurasp_learning} illustrates a probability calculation with the green arrows leading from the neural prediction to the answer set.
Since the neural network confidences are highest for the values 1 and 3, this particular answer set will have the highest probability.
The probability of an observation $O$ is the sum of the probabilities of all interpretations that model $O$:
\[P(O)=\sum_{I\models O}P(I)\]

\subsubsection{Gradient calculation}

The learning goal of NeurASP corresponds to finding neural network parameters that maximize the log likelihood of all observations in the dataset.
The gradient of this goal with regard to neural parameters $\bm{\theta}$ is given by:
\[\sum_O\frac{\partial\log(P(O))}{\partial\mathbf{p}}\times\frac{\partial\mathbf{p}}{\partial\bm{\theta}}\]
where $\mathbf{p}$ represents the output vector of the neural network for the example relevant to $O$.
The second part of the multiplication is processed automatically by backpropagation frameworks.
The first part is calculated explicitly for each entry of the vector, where $p_v$ denotes the vth entry of $\mathbf{p}$.

\begin{equation}
\frac{\partial\log(P(O))}{\partial p_v}=\frac{\sum\limits_{\substack{I:I\models O\\I\models c=v}}{\frac{P(I)}{P(c=v)}-\sum\limits_{\substack{I,v':I\models O\\I\models c=v',v\neq v'}}{\frac{P(I)}{P(c=v')}}}}{\sum\limits_{I:I\models O}{P(I)}}
\label{eq:grad}
\end{equation}

Intuitively, each time an answer set includes the current prediction, the gradient is increased (left-hand side) and if it includes another value for this atom, it is decreased (right-hand side).
Using gradient ascent, this calculation encourages predictions that agree with more answer sets.
Figure~\ref{fig:neurasp_learning} illustrates the two-step gradient calculation with the blue arrows leading from the answer sets to neural predictions and into the network.

\section{NeurASP improvements}
\label{sec:improvements}
While the original NeurASP paper reported good results on a variety of simple tasks~\citep{yang2020neurasp}, subsequent papers have reported timeouts and scalability issues for more complex datasets~\citep{aspis2022embed2sym,aspis2024embed2rule,cunnington2024foundation_reasoning}.
This poses an issue, as the negative results occur due to an imperfect implementation rather than theoretical limits of the framework.
In this section, we outline improvements to the implementation of NeurASP to alleviate such bottlenecks.

\subsection{Vectorization}

\begin{figure}
\begin{subfigure}{.48\linewidth}
\begin{lstlisting}[breaklines, language=Python, numbers=left, numberstyle=\tiny\color{gray}, basicstyle=\footnotesize\ttfamily,]
def prob_of_interpretation(I):
 prob = 1.0
 for ruleIdx, list_of_atoms in enumerate(prob_choices):
  for atomIdx, atom in enumerate(list_of_atoms):
   if atom in I:
    prob = prob * parameters[ruleIdx][atomIdx]
 return prob

probs = [prob_of_interpretation(I) for I in Is]
\end{lstlisting}
\caption{Original NeurASP code for the probability calculations}
\label{code:prob}
\end{subfigure}
\begin{subfigure}{.48\linewidth}
\begin{lstlisting}[breaklines, language=Python, numbers=right, numberstyle=\tiny\color{gray}, basicstyle=\footnotesize\ttfamily,]
def prob_of_interpretation_new(Is):
 net_confs = torch.stack(parameters)
 concept_indices = torch.arange(len(net_confs)).repeat(len(Is), 1)
 probs = net_confs[concept_indices, Is].prod(1)
 return probs
\end{lstlisting}
\caption{Vectorized code for the probability calculations}
\label{code:new_prob}
\end{subfigure}
\caption{NeurASP probability calculation function}
\end{figure}

NeurASP spends a considerable amount of time on calculating answer set probabilities (Equation~\ref{eq:prob}) and gradients (Equation~\ref{eq:grad}).
These calculations are slow due to the use of nested loops.
We provide an illustrative example for the probability calculation in Figure~\ref{code:prob}.
The original NeurASP code iterates through each interpretation $I$ in a list comprehension, calling \texttt{prob\_of\_interpretation}.
That function consists of two for-loops.
The outer loop iterates through the entries $c=v$ of the answer set, while the inner loop iterates through all neural atoms.
The if-statement checks whether the current atom is in the interpretation $I$, which is a list of all atoms in the answer set, including non-neural atoms.
If the check is true, the neural net prediction is multiplied with the running probability.

Our improved implementation replaces the loops with Pytorch tensor operations, as shown in Figure~\ref{code:new_prob}.
To begin with, all interpretations are stored as a tensor of numbers, rather than strings, and only include neural atoms.
The function processes all interpretations at once, rather than looping through each $I$ individually.
Similarly, all vectors $\textbf{p}$ are stacked into a single 2D tensor.
The probability function then retrieves the probability value of each atom from this tensor using indexing and multiplies it with each answer set using the Pytorch $\texttt{prod}$ function.

Similarly, the gradient calculation for Equation~\ref{eq:grad} uses four nested loops in the original code.
A list comprehension calls the gradient function for each input $x$ and its corresponding prediction vector $\mathbf{p}$.
The gradient function first iterates through each entry $p_v$.
For each entry, a loop iterates through all answer sets $I$ for that example.
For each answer set, a loop iterates through all possible atoms until the atom $c=v$ is reached that is within the answer set.
At this point, its probability is either added or subtracted to the running sum.

Our improved code again replaces all loops with Pytorch operations.
In short, it weights all answer set entries at once by their network confidences.
Then, it multiplies each entry using a tensor of 1/-1 entries depending on whether their value is present in corresponding the answer set.
Last, the entries are summed together using Pytorch's $\texttt{sum}$ function to arrive at the gradients.

\subsection{Answer set caching}

Answer sets are computed using the third-party library Clingo, leaving little room for efficiency improvements.
Instead, we avoid as many Clingo calls as possible by making use of caching.
The original NeurASP implementation already has a version of caching built in, but it is suboptimal.
Given a dataset, NeurASP calculates the answer sets of each example once and then stores them.
After the first epoch, the answer sets for the same example can be retrieved from storage.
However, this still results in a Clingo call of every dataset entry, and most datasets include 10s of thousands of examples.

We make use of the observation that the answer sets of an example only depend on its downstream label, not its inputs, as Figure~\ref{fig:neurasp_learning} illustrates.
Thus, we only have to calculate answer sets once per unique label, rather than once per example.
This leads to significantly fewer Clingo calls and substantially reduces the size of the answer set cache.
In the case of MNIST addition, the dataset contains 30,000 entries, but only 19 unique labels (all possible two-digit sums, i.e.\ the numbers 0 to 18).
This allows us to skip 29,981 out of 30,000 Clingo calls!
We also store the answer sets in a file, so that we can reuse the cache for subsequent experiments with the same dataset.

\subsection{Other improvements}
We have introduced further improvements to the NeurASP code that enhance the experience of running experiments on larger datasets.
\begin{itemize}
    \item \textbf{Batch processing:} The original code requires two separate lists for inputs and observations.
    This setup requires you to load all input data into the list first, which is very memory-intensive.
    Moreover, it processes each entry one at a time, which is slow.
    We have refactored the code to support batched data from data loaders, which can load data on the fly and run it through the neural network in batches.
    \item \textbf{Validation sets:} We include the ability to use validation sets for hyperparameter tuning.
    \item \textbf{Seeding:} To ensure reproducible results, you can provide seeds to fix the random number generators.
    \item \textbf{Logging:} Instead of just printing accuracy results, we have added more extensive logging capabilities.
    The code calculates latent and downstream accuracies every specified step and stores them in JSON files.
    Model parameters with the best accuracy are saved in a file as well.
    \item \textbf{Unit tests:} We have written unit tests for every function that we changed, ensuring that the new implementation is functionality correct.
\end{itemize}

Overall, these additions modernize the code and make it usable for extensive experimentation with large datasets.

\section{Synthetic speed tests}
\label{sec:performance}
This section compares the computation times of the gradient and probability calculations on synthetic data characterized by the number of answer sets.
We compare the times of the original and improved NeurASP implementations, as well as the SLASH framework~\citep{skryagin2022slash}, which is an extension of NeurASP and supports the same functionality.
All experiments were run on a 2x 28-core Intel Xeon Gold 6348 CPU with an NVIDIA AD102GL [RTX 6000 Ada Generation] GPU running Ubuntu.

The synthetic data consists of randomly generated values for the network predictions and answer sets.
We set the number of inputs to 10 and the number of concepts to 50, which are realistic values when considering existing datasets.
Using higher numbers has lead to probabilities rounding down to 0 in our testing, making the calculations trivial.
The number of answer sets forms the main point of variation between tasks.
Therefore, we ran our experiment for different numbers of answer sets, between one hundred and one million, to study their effect on execution times.

\begin{figure}
\includegraphics[width=\textwidth]{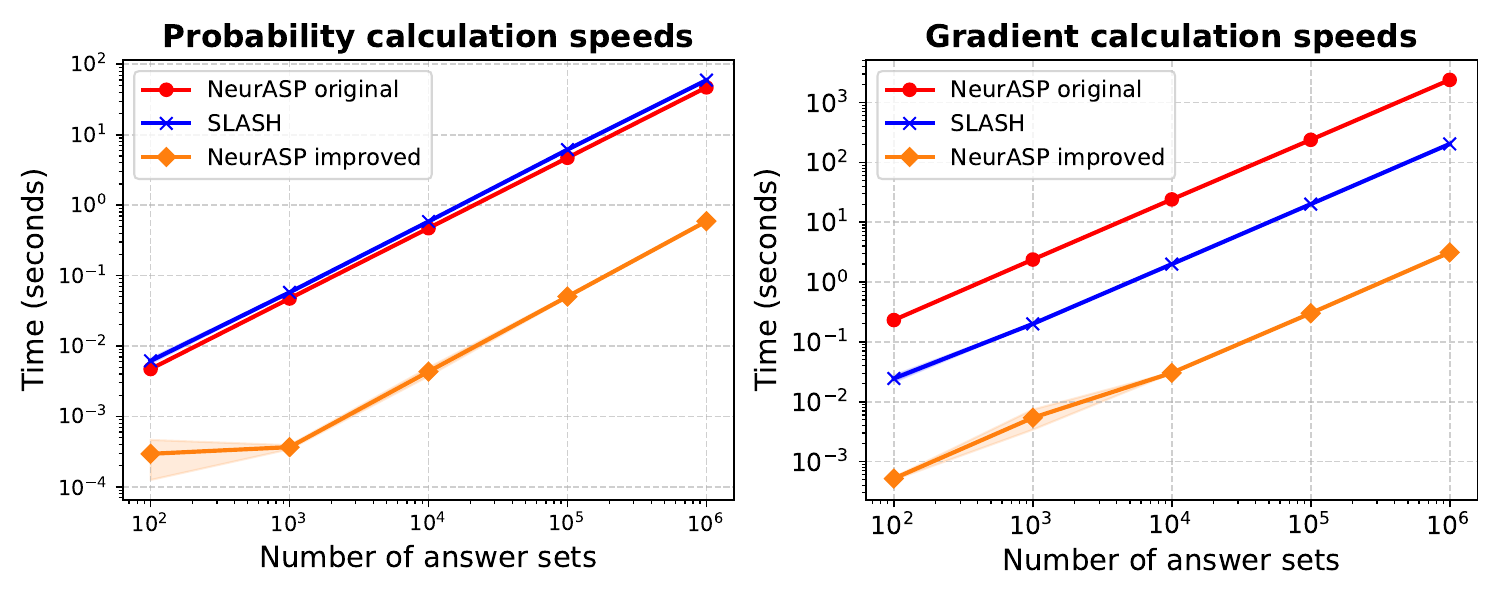}
\caption{The computation times (seconds) for the probability and gradient calculations on synthetic data, comparing the original and improved NeurASP, as well as SLASH. Each point represents the average value over five runs and the shaded area represents the standard deviation. Note the logarithmic scale on both axes.}
\label{fig:synthetic_timings}
\end{figure}

The results of our synthetic speed tests are shown in Figure~\ref{fig:synthetic_timings}.
Our improved NeurASP implementation is consistently two orders of magnitude faster than the original for probability calculations, and three orders of magnitude for gradient calculations.
For example, NeurASP used to take 0.4 seconds to calculate the probabilities of 10,000 stable models and 24 seconds for their gradients.
With the new implementation, it takes 0.004 and 0.03 seconds respectively.
While SLASH takes as much time as the original NeurASP on probability calculations, it is an order of magnitude faster for gradients.
This is because it makes use of vectorized Pytorch \texttt{einsum} functions for the gradients.
However, it still loops through splits of 10 answer sets at a time, limiting its scalability.
Therefore, our implementation still improves upon SLASH by 100x times on average.

Overall, the synthetic tests show significant scalability improvements compared to both existing frameworks.
Since these tests look at the functions in isolation, we introduce benchmark datasets in the next section, which we use for full-scale experiments.

\section{Benchmark datasets}
\label{sec:dataset}
A variety of benchmark datasets exist for neurosymbolic reasoning, but they contain simple inputs with a limited number of concepts.
We describe two of these standard tasks in this section, both of which use MNIST images and have 10 concepts.
To address the lack of challenging benchmarks, we introduce Card arithmetic, a new set of tasks with more difficult perception inputs and a larger latent space.

\subsection{MNIST addition and Member}
Both datasets use MNIST images as their inputs, which are 28x28 greyscale images of handwritten digits~\citep{deng2012mnist}.
In MNIST addition, the reasoning task consists of calculating the sum of the two input digits.
In Member, a symbolic number is given with a list of MNIST images.
The reasoning task consists of checking whether the symbolic number is contained in this list.
The downstream label is 1 if the digit is contained in the list and 0 otherwise.
Member can be configured by changing the number of images in the list, we investigate lengths of 3 and 5.

\subsection{Card arithmetic}

In Card arithmetic, the input consists of real photos of playing cards with a resolution of 523x831 pixels, which we have taken from~\citet{cunnington2023ffnsl}.
Just as in the original paper, our dataloader resizes the images to 174x274 pixels.
There are 52 latent concepts, one for every combination of rank (2-10, Jack, Queen, King, Ace) and suit (Diamonds, Clubs, Spades, Hearts).
Both the resolution and number of concepts represent a significant increase in challenge compared to the MNIST (28x28, 10 concepts) or CIFAR10 (32x32, 10 concepts)~\citep{krizhevsky2009cifar10} datasets commonly used in the literature.

The reasoning component can be freely configured to create problems with different levels of complexity.
For this paper, we have created two tasks with three levels of difficulty each.
Both tasks assign each rank a value from 2--14  and calculate the downstream result by summing up card values:

\begin{verbatim}
sum(1,V) :- card_value(1,V).
sum(P,V1+V2) :- sum(P-1,V1), card_value(P,V2).
result(X) :- sum(num_cards,X).
\end{verbatim}

The Card sum task assigns the values 0, 13, 26, and 39 to the suits and computes the \texttt{card\_value} as the sum of its rank and suit value.
The Card prodsum task assigns the values 1, 2, 3 and 4 to the suits and computes the \texttt{card\_value} as the product of its rank and suit value.

\subsection{Dataset complexities}

\begin{table}
\centering
\caption{Complexity of different tasks in terms of the average number of answer sets, the number of distinct downstream labels, the number of input images per example and the number of latent concepts per input.}
\label{tab:task_metadata}
\begin{tabular}{l cccc}
\topline
\textbf{Task} & \textbf{Answer sets} & \textbf{Labels} & \textbf{Inputs} & \textbf{Concepts} \\
\midrule
MNIST addition & 5 & 19 & 2 & 10 \\
\midrule
Member 3 & 30,000 & 20 & 3 & 10 \\
Member 5 & 50,000 & 20 & 5 & 10 \\
\midrule
Card sum 2 & 27 & 101 & 2 & 52 \\
Card sum 3 & 950 & 154 & 3 & 52 \\
Card sum 4 & 40,130 & 182 & 4 & 52 \\
\midrule
Card prodsum 2 & 28 & 95 & 2 & 52 \\
Card prodsum 3 & 1,048 & 134 & 3 & 52 \\
Card prodsum 4 & 58,580 & 174 & 4 & 52
\botline
\end{tabular}
\end{table}

Table~\ref{tab:task_metadata} provides an overview of the complexity of each task in terms of different properties.
Both Member and Card tasks reach tens of thousands of answer sets, but they differ in the number of labels and concepts.
Member always has 20 distinct downstream labels, one for each combination of label (true or false) and symbolic input (0 to 9).
The Card tasks include distinct labels for each possible result of the arithmetic operation, and their number increases with the number of inputs.
While MNIST digits include ten concepts, there are 52 different playing cards.

\section{Experiments}
\label{sec:experiments}
We analyse the performance of the improved NeurASP implementation on benchmark datasets.
In the frst section, we measure the execution times of the NeurASP versions on learning each task from scratch.
In the second section, we compare the performance of NeurASP with the neurosymbolic reasoning framework Embed2Sym.

\subsection{Benchmark speed tests}
The actual running time of NeurASP depends on more factors than just probability and gradient calculations, such as data loading, processing and answer set computations.
This section summarizes the execution times of the original and improved NeurASP, as well as SLASH, on real datasets.
Each task was run for the number of epochs necessary to achieve at least 90\% accuracy, emulating a real training run-through.
For the hardest task, we had to extrapolate the times for the original NeurASP and SLASH, as they would have taken days or weeks.

\begin{table}
\centering
\small
\caption{Average execution times (seconds). Bold values indicate the fastest time. * Times with an asterisk were extrapolated as follows: We ran the framework for three epochs and summed together the first epoch statistics with the average of the second two epochs multiplied by the number necessary to match the final number of epochs. $^{\dagger}$ For times with a dagger, both NeurASP and SLASH took multiple hours to process a few hundred datapoints. Therefore, no meaningful extrapolation of individual times was possible.}
\label{tab:timings}
\begin{tabular}{l l c @{\hspace{1em}} ccc}
\topline
\multirow{2}{*}{\textbf{Task}} & \multirow{2}{*}{\textbf{Algorithm}} & \textbf{Total} & \multicolumn{3}{c}{\textbf{Time breakdown (s)}} \\
\cmidrule(l){4-6}
& & \textbf{time} & \textbf{Model} & \textbf{Prob} & \textbf{Grad} \\
\midrule
\multirow{3}{*}{MNIST addition} & NeurASP original & 10m6s & 55 & \textbf{0.5} & 4 \\
 & SLASH & 50s & 26 & 0.9 & 8 \\
 & NeurASP improved & \textbf{27s} & \textbf{0.03} & 1 & \textbf{3} \\
\midrule
\multirow{3}{*}{Member 3} & NeurASP original & 2m7s & 50 & 4 & 33 \\
 & SLASH & 2m20s & 37 & 20 & 76 \\
 & NeurASP improved & \textbf{30s} & \textbf{0.4} & \textbf{0.3} & \textbf{5.2} \\
\cline{2-6}
\multirow{3}{*}{Member 5} & NeurASP original & 4h12m39s & 8587 & 632 & 5879 \\
 & SLASH & 6h22m22s & 14077 & 4126 & 4256 \\
 & NeurASP improved & \textbf{12m12s} & \textbf{319} & \textbf{1} & \textbf{407} \\
\midrule
\multirow{3}{*}{Card sum 2} & NeurASP original & 46m10s & 225 & 23 & 1044 \\
 & SLASH & 30m14s & 99 & 2 & 9 \\
 & NeurASP improved & \textbf{5m45s} & \textbf{0.7} & \textbf{0.6} & \textbf{3} \\
\cline{2-6}
\multirow{3}{*}{Card sum 3} & NeurASP original* & 3d21h12m & 2540 & 6350 & 318650 \\
 & SLASH & 10h20m10s & 15240 & 1200 & 4820 \\
 & NeurASP improved & \textbf{2h5m30s} & \textbf{16} & \textbf{33} & \textbf{405} \\
\cline{2-6}
\multirow{3}{*}{Card sum 4} & NeurASP original$^{\dagger}$ & $>$100d & - & - & - \\
 & SLASH$^{\dagger}$ & $>$30d & - & - & - \\
 & NeurASP improved & \textbf{17h51m40s} & \textbf{1886} & \textbf{485} & \textbf{19700}
\botline
\end{tabular}
\end{table}

Table~\ref{tab:timings} presents the results of our speed tests.
Our implementation exhibits substantially lower execution times than both the original NeurASP and SLASH\@.
The harder the tasks become, the larger the gap grows.
The time breakdown illustrates which enhancements save the most time.

The answer set caching results in the largest overall speed up.
NeurASP calls Clingo 30,000 times for MNIST addition and 1,000 times for Card sum and Member before it makes use of its cache in epochs 2 and 3.
SLASH does not cache answer sets at all, resulting in 90,000 and 3,000 calls respectively.
It does employ the \textit{SAME} method~\citep{skryagin2024scalable_slash} to reduce the size of choice rules as training goes on.
While this speeds up answer set computations, it is still slower than caching.
In contrast, our improved code only calls Clingo once per unique label, i.e.\ 19, 154 and 20 times, leading to vast improvements in model computation speed.

Probability and gradient functions also take up much less time thanks to vectorization.
Only for the easiest task, MNIST addition, is the overhead of loading data into tensors slower than NeurASP\@.
The larger the dataset is, the more substantial the time improvements get.

We have shown that our improvements do not just exist in isolation, but significantly speed up training on real datasets.
For the largest tasks, runtimes reduce from months to hours, making them feasible to learn.

\subsection{Performance tests}
The improved implementation opens up the possibility of running NeurASP on tasks it previously timed out on.
In particular, NeurASP was reported to time out on Member 5~\citep{aspis2022embed2sym} and the Card arithmetic tasks are even more challenging.
We compare the performance of our improved NeurASP implementation with the neurosymbolic framework Embed2Sym.
The experimental setup and instructions on how to run them can be found in the supplementary material.

\begin{table}
\centering
\small
\caption{Test accuracies for various tasks. Values indicate downstream and latent accuracy averages and standard deviations over five runs. We compare our improved NeurASP implementation and Embed2Sym. T/O indicates a time-out after 24 hours.}
\label{tab:performance}
\begin{tabular}{lcccc}
\topline
\textbf{Task} & \multicolumn{2}{c}{\textbf{NeurASP}} & \multicolumn{2}{c}{\textbf{Embed2Sym}}\\
& Downstream & Latent & Downstream & Latent \\
\midrule
MNIST addition & $97\pm0.1$ & $98\pm0.1$ & $97\pm0.2$ & $99\pm0.1$ \\
\midrule
Member 3  & $98\pm0.3$ & $96\pm0.5$ & $98\pm0.3$ & $95\pm1.3$ \\
Member 5  & $98\pm0.7$ & $96\pm0.9$ & $97\pm0.8$ & $93\pm1.3$ \\
\midrule
Card sum 2 &  $97\pm1.8$ & $98\pm0.9$ & T/O & T/O \\
Card sum 3 &  $83\pm6.7$ & $94\pm2.6$ & T/O & T/O \\
Card sum 4 &  $60\pm9.1$ & $86\pm4.3$ & T/O & T/O \\
\midrule
Card prodsum 2 & $98\pm0.7$ & $67\pm4.5$ & T/O & T/O \\
Card prodsum 3 & $90\pm5.4$ & $66\pm2.7$ & T/O & T/O \\
Card prodsum 4 & $74\pm10$  & $58\pm4.2$ & T/O & T/O
\botline
\end{tabular}
\end{table}

The test accuracies are shown in Table~\ref{tab:performance}.
Notably, NeurASP now matches accuracies with Embed2Sym in MNIST addition and Member, even though it timed out in the original Embed2Sym paper.
Moreover, with the Card arithmetic tasks, the opposite occurs:
NeurASP is able to finish running, while Embed2Sym times out.
The reason is that Embed2Sym clusters the latent space of the neural network and assigns symbolic labels with an ASP optimization task.
The answer set program finds assignments of clusters to latent concepts by computing whether the assignments optimize the downstream labels.
This is manageable for 10 clusters, as is the case in the MNIST-based tasks.
However, there are 52 clusters with playing cards, so the optimization task has up to $52!$ solutions, timing out after 24 hours.
Embed2Sym scales better in terms of input numbers, as it was able to solve Member 20, which NeurASP cannot solve due to millions of possible answer sets.
However, we have shown that it scales worse than NeurASP in the number of latent concepts.

Analysing the performance of NeurASP on Card arithmetic, it becomes apparent that the accuracy decreases as the number of cards increase.
For Card sum 4, each label yields more than 40k answer sets on average, yet NeurASP still manages to achieve a latent accuracy of over 80\%.
Interestingly, the standard deviation of results becomes quite large as the input size increases, reaching levels of 9--10\% for the 4-card tasks.
This is much higher than we would expect for fully neural models and showcases the volatility of semantic loss.
The network can minimize the semantic loss of a single example by matching any of its answer sets.
This leads to a large number of suboptimal local minima, since the neural network can achieve significant loss decreases by fitting to answer sets that do not match the true latent labels.

\subsubsection{Reasoning shortcuts}
Networks trained on Card prodsum exhibit signs of reasoning shortcuts, where they learn to model the downstream task with incorrect latent labels~\citep{marconato2023reasoning_shortcuts}.
As Table~\ref{tab:performance} shows, NeurASP achieves higher downstream accuracies than latent accuracies for all input sizes of Card prodsum.
The network is able to predict the correct end result with incorrect latent labels.
This is possible because different cards are assigned the same values:
For example, the 3 of Clubs has the value $3\times2=6$, which is the same as the 2 of Spades with $2\times3=6$.
The network either learns to assign these labels arbitrarily, or collapses two card labels with the same value into one.
Such reasoning shortcuts do not occur with Card sum, as each suit/rank combination has a unique value.

\begin{figure}
\begin{subfigure}{.48\linewidth}
\includegraphics[width=\textwidth]{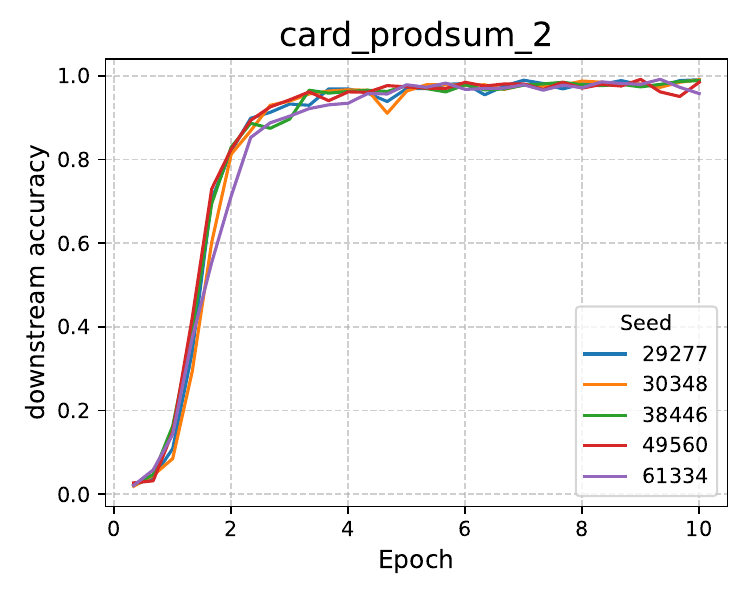}
\end{subfigure}
\begin{subfigure}{.48\linewidth}
\includegraphics[width=\textwidth]{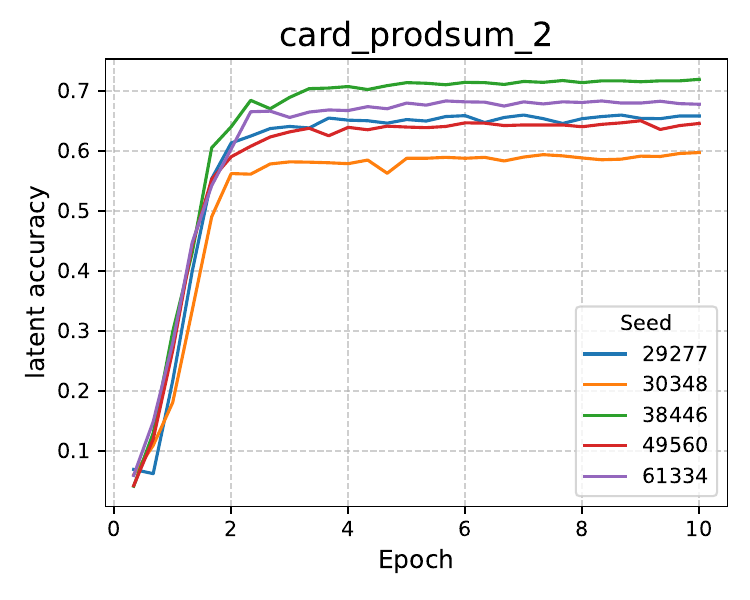}
\end{subfigure}
\caption{Differences between downstream and latent validation accuracies for Card prodsum 2 over 5 runs.}
    \label{fig:prodsum_2}
\end{figure}

Figure~\ref{fig:prodsum_2} illustrates the reasoning shortcuts for Card prodsum 2.
Even though the downstream accuracies all converge to the same value, the latent accuracies settle at vastly different values.
For each run, the network learns different assignments to latent concepts and collapses a different number of them.
Card prodsum is therefore a good dataset for investigating reasoning shortcuts and possible mitigations in the future.

\section{Conclusion}
\label{sec:conclusion}
In this paper, we implemented a variety of improvements to the NeurASP framework, facilitating its usability for research and speeding up its execution time from hours to minutes on complex neurosymbolic tasks.
With the improved framework, we stepped beyond the MNIST and CIFAR datasets typically seen in this research field and ran it on increasingly challenging tasks using playing cards as inputs.
Not only is the improved NeurASP framework able to run tasks with 10s of thousands of answer sets, its semantic loss learning function achieves high accuracies.
It outperforms the neurosymbolic framework Embed2Sym when the number of latent concepts increases.

Our improvements pave the way for future work to investigate the strengths and limits of NeurASP and semantic loss.
With the card arithmetic dataset, we have created a basis for a variety of tasks with different complexities that feature a latent space of 52 different playing cards.
A clear next step is to compare NeurASP to other neurosymbolic frameworks on this dataset and beyond.
Another interesting avenue is the investigation of reasoning shortcuts, where incorrect latent predictions can still incur a correct downstream prediction.
Techniques like regularization have shown promising results to mitigate shortcuts and could be applied to NeurASP as well.

\section*{Acknowledgements}
This work was supported by the UKRI Centre for Doctoral Training in Safe and Trusted Artificial Intelligence [EP/S0233356/1].

\bibliographystyle{tlplike}
\bibliography{bibliography}

\newpage
\appendix

\section{Code base}
The implementation of the improved NeurASP code and the Card arithmetic dataset can be found on \url{https://github.com/azreasoners/NeurASP}.
The speed comparisons with the original NeurASP implementations and SLASH are available on \url{https://github.com/APRader/NeurASP/tree/speed_tests}.
 It includes comparison functions for in \texttt{tests/speed\_tests.py}, as well as unit tests to verify the correctness of the new implementation in \texttt{tests/test\_NeurASP.py}.

\section{Experiments}
We provide the learning curves for the Card arithmetic experiments.
For all runs, we set the batch size to 64, the learning rate to 0.0001, the weight decay to 1e-5 and the checkpoint frequency to 52.
For the 2, 3 and 4 input variations, we set the number of epochs to 10, 15 and 20 respectively.

\begin{figure}
\begin{subfigure}{.48\linewidth}
\includegraphics[width=\textwidth]{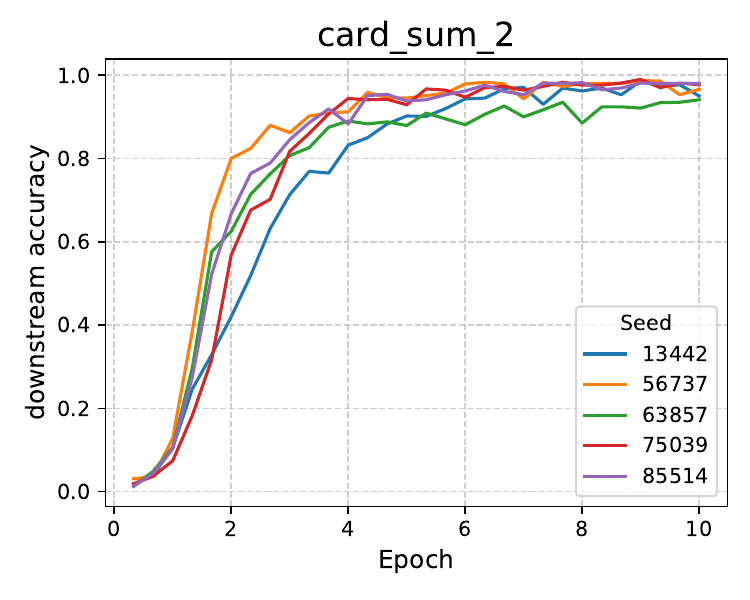}
\end{subfigure}
\begin{subfigure}{.48\linewidth}
\includegraphics[width=\textwidth]{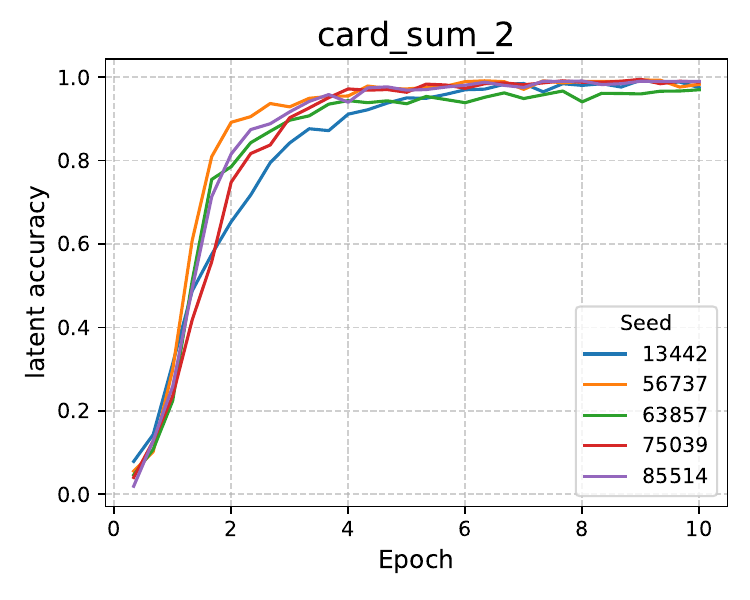}
\end{subfigure}
\caption{Downstream and latent validation learning curves for Card sum 2 over 5 runs.}
\end{figure}

\begin{figure}
\begin{subfigure}{.48\linewidth}
\includegraphics[width=\textwidth]{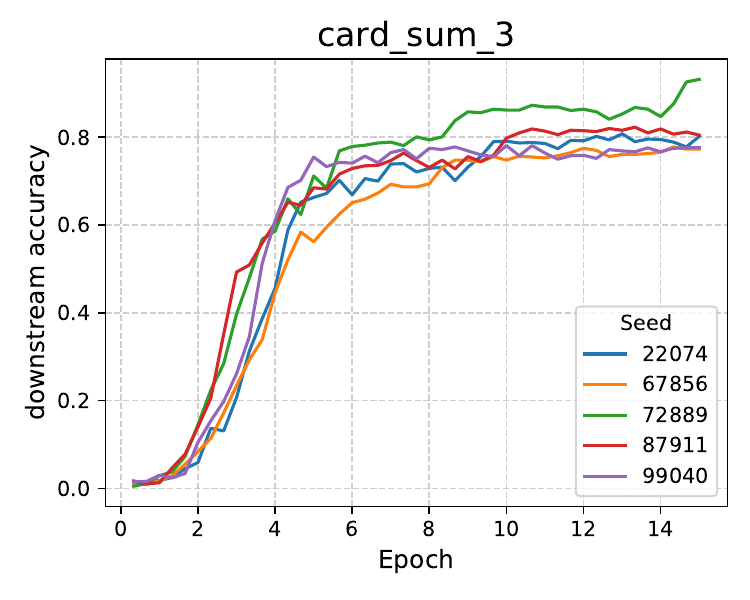}
\end{subfigure}
\begin{subfigure}{.48\linewidth}
\includegraphics[width=\textwidth]{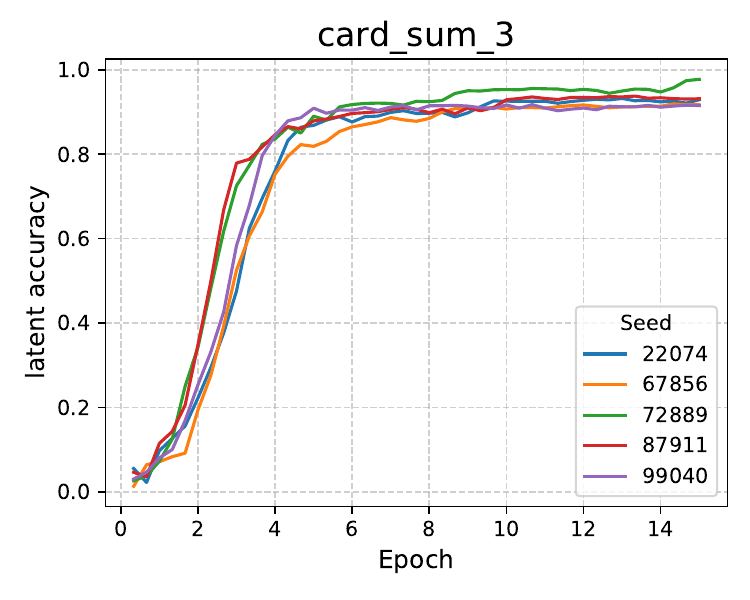}
\end{subfigure}
\caption{Downstream and latent validation learning curves for Card sum 3 over 5 runs.}
\end{figure}

\begin{figure}
\begin{subfigure}{.48\linewidth}
\includegraphics[width=\textwidth]{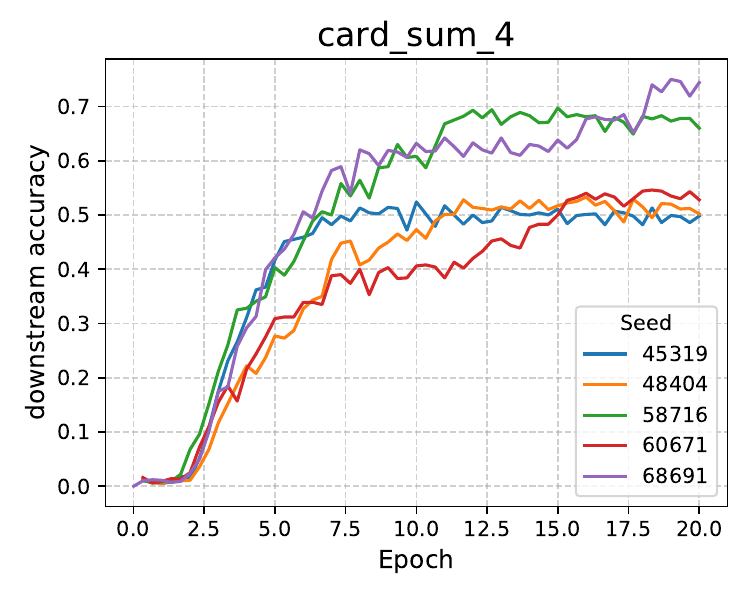}
\end{subfigure}
\begin{subfigure}{.48\linewidth}
\includegraphics[width=\textwidth]{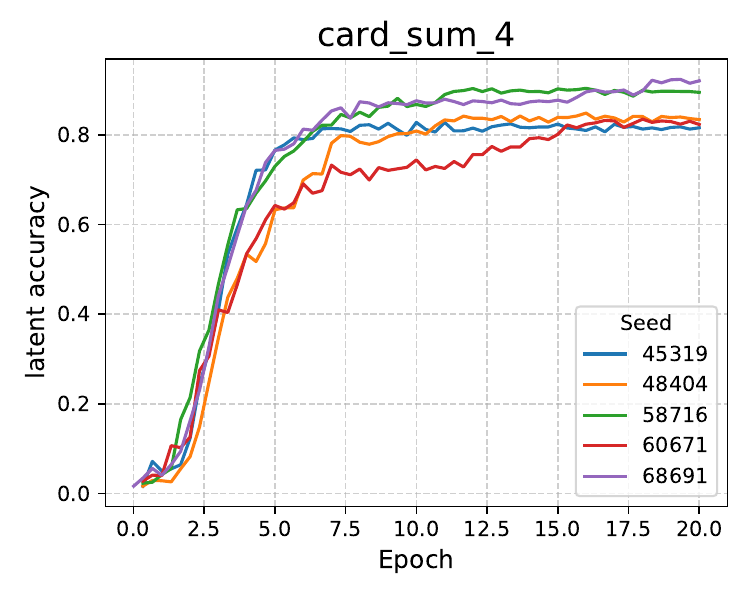}
\end{subfigure}
\caption{Downstream and latent validation learning curves for Card sum 4 over 5 runs.}
\end{figure}

\begin{figure}
\begin{subfigure}{.48\linewidth}
\includegraphics[width=\textwidth]{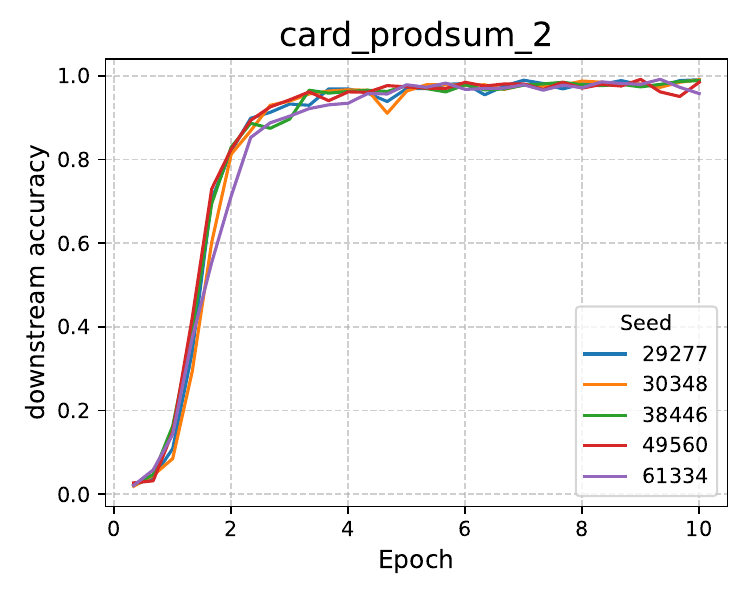}
\end{subfigure}
\begin{subfigure}{.48\linewidth}
\includegraphics[width=\textwidth]{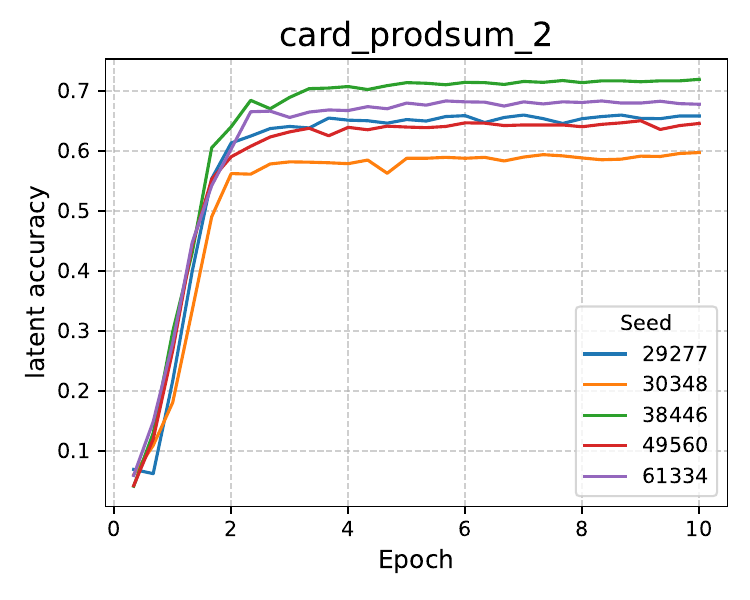}
\end{subfigure}
\caption{Downstream and latent validation learning curves for Card prodsum 2 over 5 runs.}
\end{figure}

\begin{figure}
\begin{subfigure}{.48\linewidth}
\includegraphics[width=\textwidth]{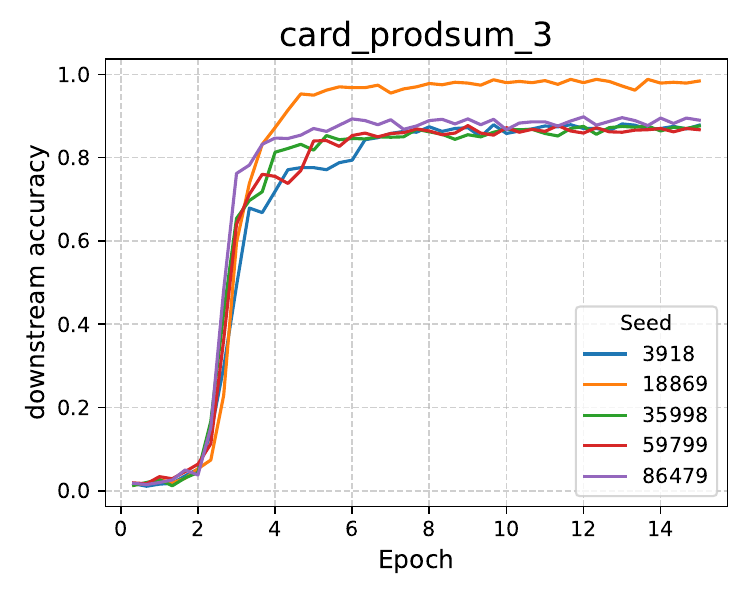}
\end{subfigure}
\begin{subfigure}{.48\linewidth}
\includegraphics[width=\textwidth]{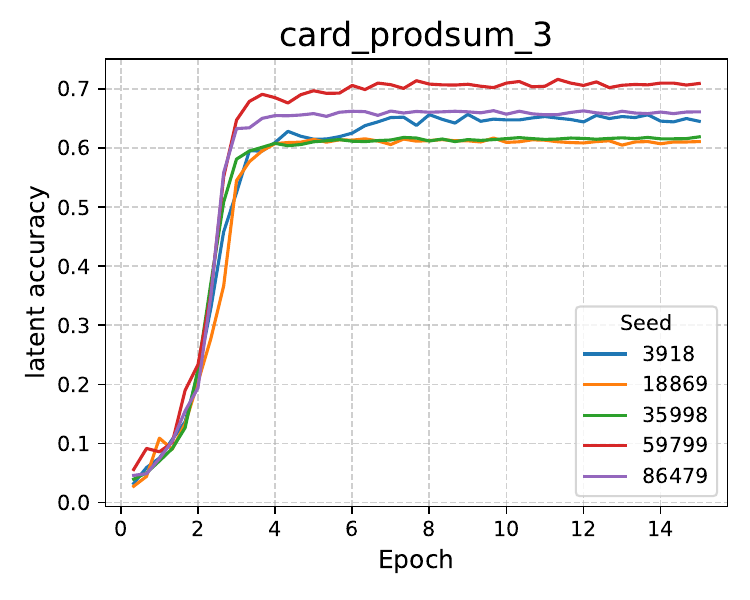}
\end{subfigure}
\caption{Downstream and latent validation learning curves for Card prodsum 3 over 5 runs.}
\end{figure}

\begin{figure}
\begin{subfigure}{.48\linewidth}
\includegraphics[width=\textwidth]{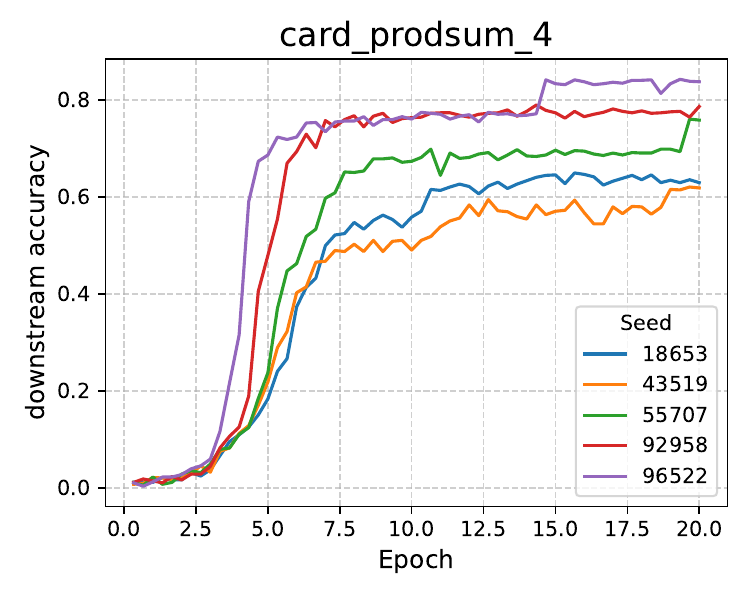}
\end{subfigure}
\begin{subfigure}{.48\linewidth}
\includegraphics[width=\textwidth]{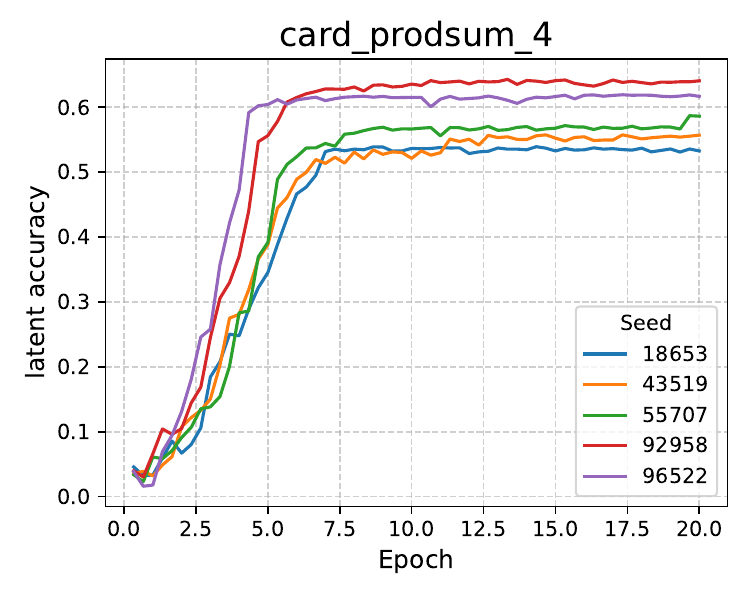}
\end{subfigure}
\caption{Downstream and latent validation learning curves for Card prodsum 4 over 5 runs.}
\end{figure}
\end{document}